\documentclass[authoryear,a4paper,10pt,twocolumn,preprint,3p]{elsarticle}

\usepackage{amsmath}
\usepackage{algorithm}
\usepackage[noend]{algpseudocode}

\usepackage{booktabs}
\usepackage{multirow}

\usepackage{caption}

\usepackage[normalem]{ulem}
\useunder{\uline}{\ul}{}

\usepackage{float}

\begin{document}
\begin{frontmatter}
\title{A heuristic scheme for the Cooperative Team Orienteering Problem with Time Windows}
\author[]{Iman Roozbeh\corref{cor1}}
\ead{iman.roozbeh@rmit.edu.au}
\author[]{Melih Ozlen}
\ead{melih.ozlen@rmit.edu.au}
\author[]{John W. Hearne}
\ead{john.hearne@rmit.edu.au}
\cortext[cor1]{Corresponding author}
\address{School of Science, RMIT University, Melbourne, Australia}

%	\newpage
%	\section{Abstract}
	
%	\begin{multicols}{2}
%%%%%%%%%%%%%%%%%%%%%%%%%%%%%%%%%%%%%%%%%%%%%%%%%%%%%%%%%%%%%%%%%%%%%%%%%%%%%%%%%%%%%%%%%%%%
\begin{abstract} 
 The Cooperative Orienteering Problem with Time Windows (COPTW)is a class of problems with some important applications and yet has received relatively little attention. In the COPTW a certain number of team members are required to collect the associated reward from each customer simultaneously and cooperatively. This  requirement to have one or more team members simultaneously available at a vertex to collect the reward, poses a challenging OR task. Exact methods are not able to handle large scale instances of the COPTW and no heuristic schemes have been developed for this problem so far. In this paper, a new modification to the classical Clarke and Wright saving heuristic is proposed to handle this problem. A new benchmark set generated by adding the resource requirement attribute to the existing benchmarks. The heuristic algorithm followed by boosting operators achieves optimal solutions for 64.5\% of instances for which the optimal results are known. The proposed solution approach attains an optimality gap of 2.61\% for the same instances and solves benchmarks with realistic size within short computational times. 
 
\end{abstract}
%%%%%%%%%%%%%%%%%%%%%%%%%%%%%%%%%%%%%%%%%%%%%%%%%%%%%%%%%%%%%%%%%%%%%%%%%%%%%%%%%%%%%%%%%%%%%%%%%%%%


%//******************************************************************************************************

\end{frontmatter}

%%%%%%%%%%%%%%%%%%%%%%%%%%%%%%%%%%%%%%%%%%%%%%%%%%%%%%%%%%%%%%%%%%%%%%%%%%%%%%%%%%%%%%%%%

%	\newpage 
	\section{Introduction}
		
	The \textit{Orienteering Problem} (OP) is a well-known integer programming problem in combinatorial optimisation, introduced by \cite{golden1987orienteering}. OP emerges from a combination between the travelling salesman problem and the knapsack problem. Therefore, it is a routing problem where travelling to all vertices is often not feasible due to limited resources and the available time. The objective of the OP is to find the combination of nodes that maximises the total rewards collected. Tourist trip design (\cite{verbeeck2014extension,vansteenwegen2007mobile,de2015multiobjective}), home fuel delivery (\cite{golden1987orienteering}) and asset protection during wildfire (\cite{van2014mixed})are some examples of application found in the literature.

	Various extensions of the OP have been inspired by real-world problems, such as the \textit{ Team Orienteering Problem with Time Window} (TOPTW) (\cite{labadie2012team,souffriau2013multiconstraint,duque2015solving,gunawan2015iterated}), time-dependent orienteering problem (\cite{abbaspour2011time,mei2016efficient,gavalas2014efficient}) and other variants (\cite{salazar2014multi,van2014mixed,varakantham2015direct}). Readers are referred to the survey articles by \cite{vansteenwegen2011orienteering,Gunawan2016315}, for a complete list of applications and extensions.
	
	The Orienteering problem is NP-hard  (\cite{golden1987orienteering}). An extensive number of exact and heuristic approaches have been proposed to solve this problem. In the literature, a few papers have focused on exact methods(\cite{keshtkaran2016enhanced,poggi2010team,dang2013branch}) but most papers deal with heuristic approaches(\cite{bouly2010memetic,liang2013multiple,marinakis2015memetic,dang2013effective,gunawan2015iterated,labadie2012team}).
	
	Given a set of nodes in a TOPTW problem a subset of vertices should be visited in order to collect corresponding rewards (\cite{vansteenwegen2011orienteering}). An extension to TOPTW is the \textit {\textbf{  Cooperative}} {\it Orienteering Problem with Time Windows} (COPTW), proposed by \cite{van2015optimisation}. This problem arises when some tasks that need to be undertaken at certain locations can only be accomplished by two or more individuals acting cooperatively.	In COPTW each node has a unique resource requirement. Collecting the corresponding reward at any individual vertex is upon the condition of meeting the requirements within its time window. Depending on the requirement of each node, one or more team members must be available there to start the service simultaneously. This study seeks to propose a heuristic approach for solving the COPTW, for the first time.
	
	In all orienteering problems with time windows the routes are very sensitive, either a minor change in service start time at any node or even a single swap in orientation of the nodes requires additional computational effort to double-check all constraints for every single route. This very attribute motivates developing a fast and efficient heuristic to solve the problem. We handled the existing complexity of the COPTW by proposing a new robust enhancement to the \textit{Clarke and Wright} (CW) algorithm followed by boosted operators. The main feature of the heuristic algorithm is its simplicity which facilitates the handling of the problem constraints.     

%	As COPTW has not been studied previously, there is no existing benchmark instances. Thereupon, we developed a new set of instances by randomly adding up the resource requirement attribute to the \cite{vansteenwegen2009iterated,montemanni2009ant} benchmark sets. Performance of the MCW algorithm compared with the achieved optimal solution by CPLEX for small-scale problems. The experimental results demonstrate the efficiency and accuracy of the designed heuristic solution. 
	
	The remainder of the paper is organised as follows: section 2 describes the mathematical formulation of the COPTW problem. In section 3 we introduce the general scheme of the proposed heuristic approach. Our findings from an extensive experimental study are discussed in section 4. Finally, the conclusion and potential future research directions are stated in section 5.

	\section{Cooperative Orienteering Problem with Time Windows}
	
	COPTW generalises the TOPTW formulation (see \cite{van2014mixed} and \cite{ van2015optimisation} for a detailed discussion), where a certain number of team members (resources) $R=\{r_1,\dots,r_n\} $ are required at vertices $V=\{v_1,\dots,v_n\} $. The location $v_i$ is considered served if $r_i$  team members  arrive at the vertex within the time window $[o_i,c_i]$ and start the service simultaneously at time $s_i$ for a duration of $a_i$ units of time. In COPTW, a homogeneous fleet of $P$ team members start their route from $v_1$ and must return to depot $v_N$ by time $T_{max}$, where both $v_1$ and $v_N$ represent a same location. For any two vertices, $t_{ij}$ indicates the required time for each team member to travel from $i$ to $j$. The binary decision variable $y_i$, takes the value $1$ if $r_i$ team members visit $v_i$ within the appropriate time window for the required duration, $0$ otherwise. The other decision variable is $z_{ij}$ which takes $1$ if arc $ij$ is traversed, otherwise $0$. Furthermore, $x_{ij}$ represents the number of team members travelling from $i$ to $j$. Lastly, $\xi_{ij}$ defines a set of arcs that can be traversed if a team member departs vertex $i$ at $o_i+a_i$ and arrives  vertex $j$ by $c_j$. Following the definition of $\xi_{ij}$, feasible arcs that can be traversed to and from node $i$ are shown by $\Omega^{-}_{i}$ and $\Omega^{+}_{i}$, respectively.

	The mathematical formulation of the COPTW as a mixed-integer program is as follows:
	
	%1
	\begin{equation}
	Maximise \sum_{i=2}^{N-1} \sigma_iy_i 
	\end{equation}
	%2
	
	\begin{equation}
	s.t.:\sum_{j \in \Omega^{+}_{1}} x_{1j} = \sum_{i \in \Omega^{-}_{N}} x_{iN}=P,
	\end{equation}
	
	%3
	\begin{equation}
	\sum_{i \in \Omega^{-}_{k}} x_{ik} = \sum_{j \in \Omega^{+}_{k}} x_{kj}, \; k=2,\dots,N-1,
	\end{equation}
	
	%4
	\begin{equation}
	r_{k}y_{k} = \sum_{j \in \Omega^{+}_{k}} x_{kj}, \; k=2,\dots,N-1,
	\end{equation}
	
	%5
	\begin{equation}
	x_{ij}\leq P z_{ij}, \;(i,j)\in \xi,
	\end{equation}
	
	%6
	\begin{equation}
	s_i+t_{ij}+a_i-s_j\leq M(1-z_{ij}), \;(i,j)\in \xi,
	\end{equation}
	
	%7
	\begin{equation}
	o_i \leq s_i, \; i=1,\dots,N,
	\end{equation}
	
	%8
	\begin{equation}
	s_i \leq c_i, \; i=1,\dots,N,
	\end{equation}

	%9
	\begin{equation}
	x_{ij} \in \{0,1,\dots,P\}, \; (i,j)\in \xi,
	\end{equation}

	%10
	\begin{equation}
	y_i,z_{ij}\in \{0,1\}, \; (i,j)\in \xi.
	\end{equation}

	The objective function (1) maximises the sum of the rewards $\sigma_i$ collected at each vertex $i$.  Constraint (2) ensures all members depart from and return to a designated depot. Constraint (3) guarantees flow conservation by enforcing the equality of incoming and outgoing arcs to each node. Constraints (4) and (5) ensure that the collection of a reward (ie a score) at each location is dependent upon the condition of fulfilling its resource requirement, and travelling members of a fleet through an arc  never exceeds the total $P$. Constraint (6) ensures that at each vertex the service can only be started when the previously visited location has been served completely and there is sufficient time to travel to the vertex. Constraints (7) and (8) enforce every vertex is visited within its time window. Integer and binary conditions are defined in constraints (9) and (10). 
	
	A graphical representation of a sample solution for COPTW is sketched in Fig. 1 as below. 

	%\newpage
	\begin{figure}[h!]
		\centering
		\includegraphics[width=0.46\textwidth]{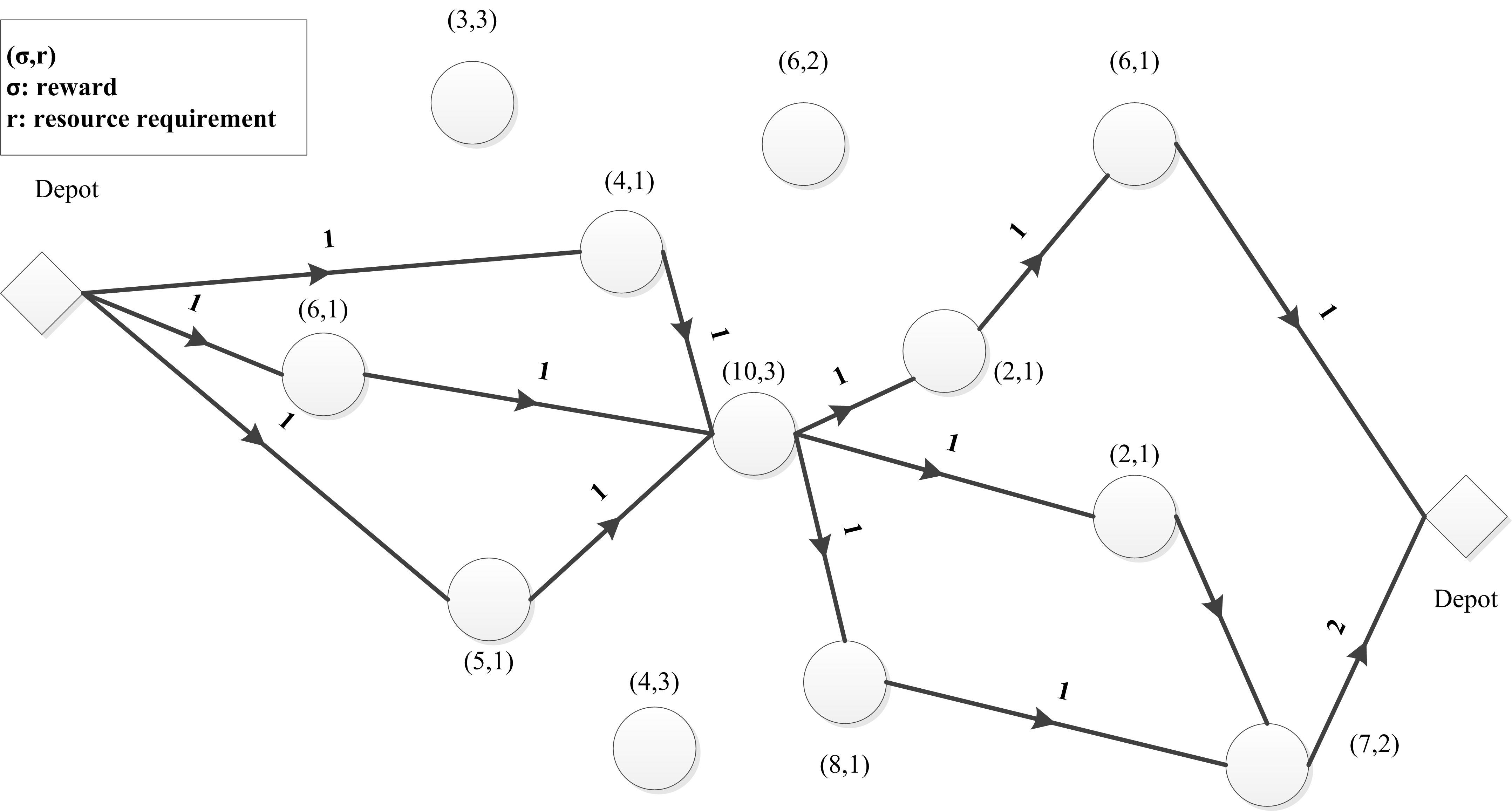}%,height=0.0\textheight
		\caption{A sample solution of the COPTW}
	\end{figure}

	Two of the main attributes of each node, the resource requirement ($r_i$) and corresponding reward ($\sigma_i$), are specified. Three team members leave the depot and ($r_i$) members must be at each vertex $i$ before starting the service to collect associated the reward. The numbers over each arc represent the travelling members of the team through an arc. It is infeasible to visit all vertices within their time frames. Considering the set of all feasible solutions a schedule is produced that maximises the score (ie the sum of rewards collected). 
		
	\section{A heuristic approach for the COPTW}
	
	The requirement in COPTW problems  that two or more members of a team are required to be at some of the locations and within specified time windows pose complications observed in neither VRP's nor  OP's.       
	
	%Service requirement of vertices and score collection upon the existence of sufficient number of team members within each time window alongside with dynamic change in service start time after each insertion or removal, pose complications that cannot be observed in neither VRPs nor OPs, or most variants thereof.
	% Thus, designing a heuristic to solve COP requires a new approach toward this problem as none of the introduced heuristics can be implemented directly in solving COPTW. The new attribute of the problem in which multiple team members are required to start their services simultaneously is a great barrier to adapt existing algorithms.
	Robust enhancements are applied to the classical Clarke and Wright Saving (CWS) heuristic (\cite{clarke1964scheduling}) followed by modified operators to tackle the challenges introduced by COPTW. The resulting algorithm is named as the \textit{Modified Clarke and Wright} (MCW) heuristic. 
	 
    %\cite{cordeau2002guide} categorised well-known classical heuristics and metaheuristics for vehicle routing problem. Considering the fact that solution quality of metaheuristic algorithms are at cost of computation time, and most importantly the problem-specific attributes which demand an efficient algorithm in term of speed, we focused on improving the accuracy of CW saving heuristic without harming its simplicity and speed which are superior compare to other algorithms. 
	
	While COPTW aims to maximise the collected score by meeting the resource requirements of vertices within their associated time windows, this can be achieved along with construction of the shortest possible routes, at least in most cases. In the proposed saving function the first two terms are inspired by extensive studies in this area. particularly, \cite{doyuran2011robust} introduced a new saving function which not only covers the accomplished improvements by \cite{altinel2005new} but also integrates a sweep algorithm for further enhancements. However, the last term is developed to explain the orienteering problem attribute to maximise the total collected score. The first term of the saving function improves the reshaping ability of the classical CW heuristic and its circumference characteristic. Motivation of the second term is to give early placement to pairs in vicinity of the depot by including $\cos\theta_{ij}$, which is the value of constructed angles between pairs. Therefore, pairs with acute angle climb in the saving pair list while the ones with absolute angles get penalise. The last term is award-based, as pairs with higher score should be at high priorities in the saving pair list, with less impression from other factors. The proposed saving function for the MCW heuristic is therefore as follows: 
	%In the first two terms $d^{max}$ is the maximum distance and $d_{ij}$ represent the distance between vertex $i$ and $j$.
	
	\begin{equation}
	\begin{gathered}
	S_{i,j}=\dfrac{d_{i0}+d_{0j}-\lambda d_{ij}}{d^{max}}+\\
	\mu\ast\dfrac{\cos \theta_{ij} | d^{max}-(d_{i0}-d_{0j})/2|}{d^{max}}+
	\vartheta\ast \dfrac{\Gamma_{i}+\Gamma_{j}}{\overline{\Gamma}} 
	\hspace{-25pt} 
	%\raisetag{40 \baselineskip}
	\rule[0em]{0pt}{20pt}
	\end{gathered}
    \end{equation}

	Where $\theta_{ij}$ represents the angle between customer $i$ and $j$ resulted by two rays originating from the depot and crossing them. In the above formulation,$S_{i,j}$, $d$ and $\Gamma$ demonstrate saving value, distance and score, respectively. In equation (11), $d^{max}$ is the maximum distance and $\overline{\Gamma}$ is the average score for each node. Moreover, three parameters namely $\lambda$, $\mu$ and $\vartheta$ are finely tuned to equip the MCW heuristic to perturbation move. To ensure the assignment of nodes with highest saving values prior to others, we implemented parallel route construction in the MCW algorithm which fit the best with underlying nature of the problem.
	
	In the well-known classical CW saving heuristic \cite{clarke1964scheduling}, all vertices are initially connected to the depot. then, saving values of pairs are obtained by joining each of the two nodes in succession. The procedure adds a new vertex at a time to the ends of tours. But, the MCW heuristic is equipped with more than one insertion operator. Covering vertices in the MCW begins by checking the ends of the constructed tours for insertion. In case of failure, the new vertices will be added within tours, as inspired by the cheapest insertion algorithm (see \cite{rosenkrantz1974approximate}) in accordance with the associated saving values. It is worthwhile to mention that none of the constraints should be violated and no removal is allowed before starting the $improve ()$ function. In the other words, the algorithm is equipped with the same logic as exists in the ejection chains (\cite{glover1996ejection}).
	
		\begin{figure}[h!]
			\centering
			\includegraphics[width=0.46\textwidth]{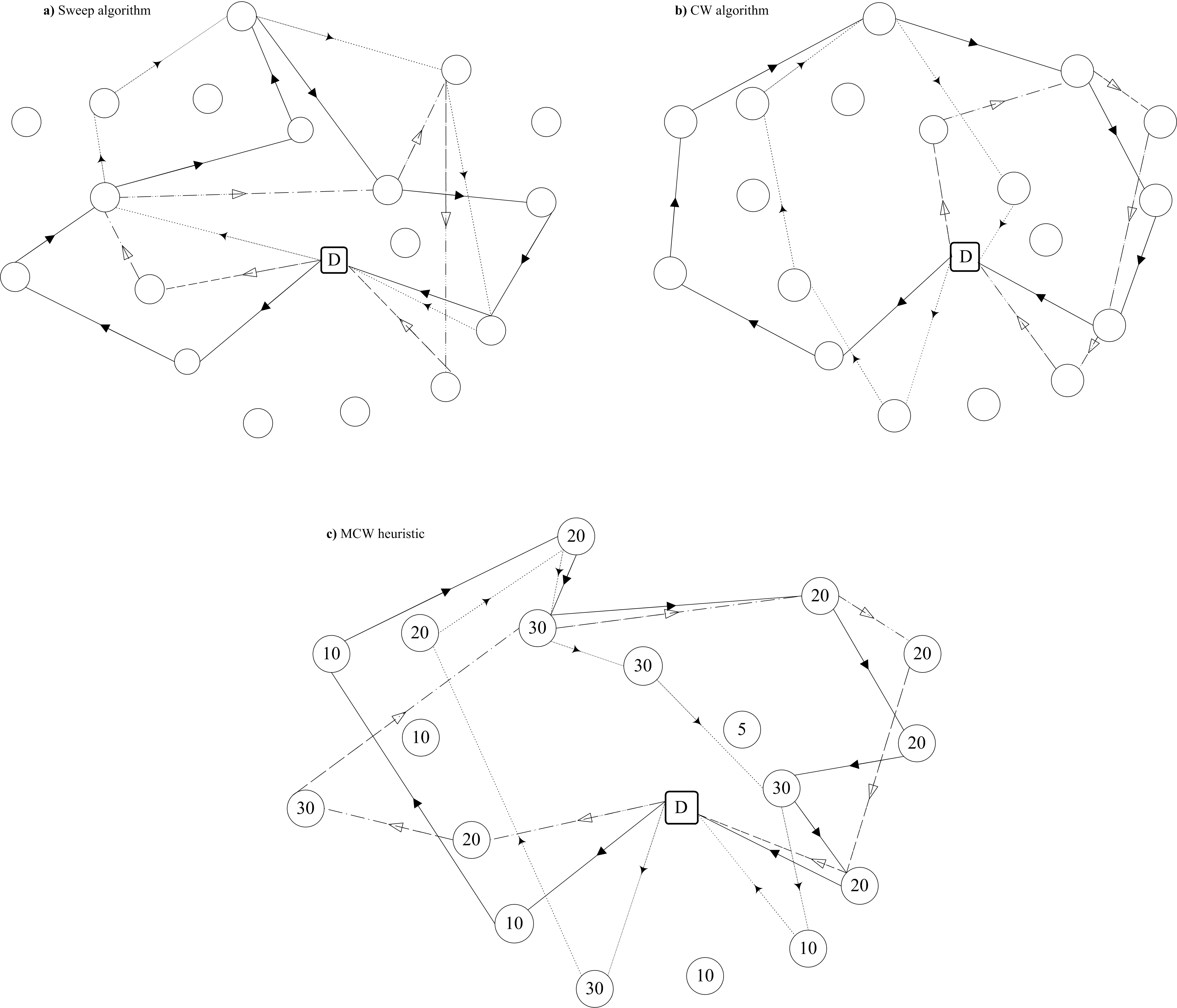}%,height=0.0\textheight
			\caption{Comparison of the routes formed by three approaches for COPTW}
		\end{figure}
	Figure 2 illustrates the effect of taking advantage from the sweep algorithm and the saving algorithm in the MCW heuristic. The depot is denoted as "D" and associated rewards are defined in figure 2(c). The CW heuristic constructs circumferenced routes (figure 2(b)) while the sweep algorithm forms routes based on polar angles with the depot regardless of distance between any pairs(figure 2(c)). The proposed saving function reshapes the constructed routes in figure 2(a) by integrating the cosine value. As can be observed nodes with higher rewards are covered in figure 2(c), which proves the importance of the last term in equation 11. The existing routes in figure 2(c) can be improved after implementation of the $improve ()$ function operators.
	
	As far as the parameters in the saving function are concerned, the computational effort is positively correlated to the intervals for $(\lambda, \mu, \vartheta)$ triplets. Based on extensive tests and considering the compromise between the search effort and solution quality, promising incremental size and intervals for the coefficients are defined. It was decided to change the parameters within $[0,1.4]$ for $(\lambda, \mu)$ pairs and $[0,3.5]$ for $\vartheta$, all with an increment of $0.7$. The large interval and step-size assist the heuristic to explore a broader area of the solution space. The pseudo-code of the presented solution is given in Algorithm 1.
	%%%%%%%%%%%%%%%%%%%            ALG                  %%%%%%%%%%%%%%%%%%%%%%%%%%%%%%%%%%%%%%%%%%%%%%%%%%    1
	\makeatletter
	\def\BState{\State\hskip-\ALG@thistlm}
	\makeatother
	\begin{algorithm}
		\caption{Pseudocode for the MCW heuristic}\label{euclida}
		\begin{algorithmic}[1]
			\Function{MCW \textit{heuristic} } {temporary routes $\tau$, best routes $\beta$, collected award $\alpha$, set of all nodes $\nu$	, set of assigned nodes $\gamma$, Saving Pair List \textit{SPL}, unvisited vertices $P$, service requirement $r$}
			\For {\textbf{all} ($\mu,\, \vartheta,\, \lambda$)}
			\BState Call the {$CalcSavingPairs (i,j)$ }function
			\For{\textbf{all} ($i\in \textit{SPL}$) }
			\BState \small{assignment of feasible $(j \in P)$ to ($\textit{subroutes} \in\tau$)}
			\BState Call the { $FeasMatrix (i,j)$} function
			\BState \textit{$VisitCount_{j}$} $\gets $ \textit{$ VisitCount_{j}+1$} 
			\If {(\textit{$ VisitCount_{j}\geq$ $r_{j}$)} }
			\BState  $\alpha_{current} \gets \alpha_{current} + \Gamma_{j}$,  $\gamma \gets j $
			\EndIf %\State  \textbf{end if}
			\EndFor \State \textbf{end for}
			\For{\textbf{all} $(j\in \nu)$ }
			\If{$(VisitCout_{j}\leq$ $r_{j})$}
			\BState {remove $j$ from \textit{tours} }, update $\tau$
			\EndIf %\State  \textbf{end if}
			\EndFor \State  \textbf{end for}
			\If {($\alpha_{current}>\,\alpha_{best}$)}
			\BState	$\alpha_{best} \gets \alpha_{current}$, $\beta \gets \tau$
			\EndIf %\State  \textbf{end if}
			\BState Call the {$Improve$} function
			%	\BState  \textbf{Until} \text{stop criterion fulfilled}
			\EndFor \State  \textbf{end for}
			%	\BState \textbf{close};
			%%check after improve!!!!!!!
			\BState \textbf{Return $\alpha_{best} \,and\, \beta$};
			\EndFunction
		\end{algorithmic}
	\end{algorithm}
	%%%%%%%%%%%%%%%%%%%%%%%%%%%%%%%%%%%%%%%%%%%%%%%%%%%%%%%%%%%%%%%%%%%%%%%%%%%%%%%%%%%%%%%%%%%%%%%%%%%%%%%%%%%%%%%%%%%
	
At first, a saving pair list is initialised by calling the function $CalcSavingPairs$. after that, unvisited nodes based on the pairs that they belong to and their saving values should be assigned to sufficient number of routes. Vertices with satisfied resource requirements are marked as serviced and relevant information are updated. However, nodes with less number of visits than their resource requirements must be removed, before applying the local search to the solution.

The $ CalcSavingPairs()$ function is represented in Algorithm 2. The saving values of feasible pairs are defined in tuples. The saving function is designed based on the sweep algorithm and CW heuristic (\cite{doyuran2011robust}) where a new additional term is added to give early placement to vertices with higher scores. The coefficient interval for this term is broader compare to the other two as it plays a crucial rule in maximisation of the objective function. Since frequent update in the $feasMatrix$ is required after each insertion or removal, this procedure is described in Algorithm 3.
%%%%%%%%%%%%%%%%%%%            ALG                  %%%%%%%%%%%%%%%%%%%%%%%%%%%%%%%%%%%%%%%%%%%%%%%%%%    2
\makeatletter
\def\BState{\State\hskip-\ALG@thistlm}
\makeatother

\begin{algorithm}
	\caption{Pseudocode for the CalcSavingPairs}\label{euclidb}
	\begin{algorithmic}[1]
		\Function{CalcSavingPairs} {saving function parameters ($\vartheta,\mu,\lambda$), Saving Pair List \textit{SPL}, travel velocity $V$, distance matrix $d_{\nu\times\nu}$}
		\BState generate $d_{\nu\times\nu}$
		\BState \textbf{Initialise: } $\vartheta,\mu,\lambda$
		\For{ ($(i,j)\in \nu$) }
		\If {(($ O_{j}+a_{j}+d_{j0}/ V \leq T_{max}$) and ($ O_{i}+a_{i}+d_{ij} / V \leq C_{j}$))} 
		\BState $S_{i,j} \gets\dfrac{d_{i0}+d_{0j}-\lambda d_{ij}}{d^{max}}+\vartheta\ast \dfrac{\Gamma_{i}+\Gamma_{j}}{\overline{\Gamma}} +\mu\ast\dfrac{\cos \theta_{ij} | d^{max}-(d_{i0}-d_{0j})/2|}{d^{max}}$
		
		\BState  insert $S_{i,j}$ to a vector of tuples $(i,j,S_{i,j})$
		\EndIf %\State  \textbf{end if}
		\EndFor \State \textbf{end for}
		\BState \textbf{Return $SPL$};
		\EndFunction
	\end{algorithmic}
\end{algorithm}
%%%%%%%%%%%%%%%%%%%            ALG                  %%%%%%%%%%%%%%%%%%%%%%%%%%%%%%%%%%%%%%%%%%%%%%%%%%    3
    Insertion and removal of each vertex can trigger significant changes in the problem environment as revision of the feasibility matrix may be required. In algorithm 3, the start time of the service at each node is the maximum value of the latest arrival time of team members to a node and its opening time. The generated solution by the MCW heuristic is already promising due to the applied logic in the saving function and the insertion operators; however algorithm 4 is designed to further improve the solution quality. 
%%%%%%%%%%%%%%%%%%%%%%%%%%%%%%%%%%%%%%%%%%%%%%%%%%%%%%%%%%%%%%%%%%%%%%%%%%%%%%%%%
\makeatletter
\def\BState{\State\hskip-\ALG@thistlm}
%	\makeatother
\begin{algorithm}
	\caption{Pseudocode for the FeasMatrix }\label{euclidc}
	\begin{algorithmic}[1]
		\Function{FeasMatrix } {service start time $\varphi$, existing nodes in sub-tours $\kappa$, latest arrival time of members to a location $T$, feasibility to travel $ feas$}
		\BState \textbf{Require: }{(temporary routes $\tau$ )}
		\For{ \textbf{all} ($i\in \nu$) }
		\BState $\varphi_{i} \gets O_{i} $
		\EndFor \State \textbf{end for}
		\For{ \textbf{all} ($\kappa\in \tau$) }
		\BState compute $T$
		\BState $\varphi_{i} \gets max{ (\varphi_{i}, T)} $
		\EndFor \State \textbf{end for}
		\For{ \textbf{all} ($\kappa\in \tau$) }
		\If{$(\varphi_{\kappa}\ge C_{\kappa}$)}
		\BState $feas_{\kappa-1 , \kappa} \gets false $
		\Else 
		\BState $feas_{\kappa-1 , \kappa} \gets true $
		\EndIf %\State \textbf{end if}
		\EndFor \State \textbf{end for}
		\BState \textbf{Return $FeasMatrix$};
		\EndFunction
	\end{algorithmic}
\end{algorithm}

%%%%%%%%%%%%%%%%%%%%%%%%%%%%%%%%%%%%%%%%%%%%%%%%%%%%%%%%%%%%%%%%%%%%%%%%%%%%	

%%%%%%%%%%%%%%%%%%%            ALG                  %%%%%%%%%%%%%%%%%%%%%%%%%%%%%%%%%%%%%%%%%%%%%%%%%%    4
\makeatletter
\def\BState{\State\hskip-\ALG@thistlm}
\makeatother
\begin{algorithm}
	\caption{Pseudocode for the improve function}\label{euclid}
	\begin{algorithmic}[1]
		\Function{\textit{improve} } { unvisited vertices $P$ }
		\BState \textbf{Require: }{( best collected award $\alpha$, best routes $\beta$)}
		\For{\textbf{all} ($i\in P$) }
		\BState $ R \, \prime \gets \beta$
		\BState  Seek to insert i
		\BState Call the {$FeasMatrix\, (i,j)$} function
		\If {(insertion cause infeasibility of node j)}
		\If { ($\Gamma_{j}\le \Gamma_{i}$)}
		\BState insert i and remove j
		\BState update $\beta$ 
		\BState update $\alpha$
		\Else
		\BState $ \beta \gets R\prime$
		\EndIf
		\Else
		\BState  update $\beta$
		\BState  update $\alpha$
		\EndIf
		\BState Call the {$FeasMatrix (i,j)$} function
		\EndFor \textbf{end for}
		\State \textbf{Return $\beta$ and $\alpha$};
		\EndFunction
	\end{algorithmic}
\end{algorithm}

%%%%%%%%%%%%%%%%%%%%%%%%%%%%%%%%%%%%%%%%%%%%%%%%%%%%%%%%%%%%%%%%%%%%%%%%%%%%%%%%%%%%%%%%%%%%%%%%%%%%%%%%%%%%%%%%%%
		The solution then undergoes a local search improvement for further enhancements which is described in Algorithm \ref{euclid}. Saving values of pairs bring on assignment of vertices in an efficient manner which cause the least additional travel distance and thereby saving more time for score collection at as many locations as possible. The $improve ()$ function checks whether any of the inserted vertices can be substituted with the unvisited ones with the possibility of improving the objective function while satisfying all constraints. Finally, it is noteworthy that, while we switch the search space by changing the parameters (perturbation move), the $improve ()$ function acts as a local search algorithm.
%	\newpage
	\section{Computational results}
    %benchmark instances
	Extensive numerical studies were conducted to evaluate the efficacy of the proposed solution approach. A set of benchmark instances was generated by adding the problem-specific attribute to the well-known existing benchmark sets (see \cite{vansteenwegen2009iterated}). The resource requirement attribute was added to each vertex by picking a random number from a set of $\{1,2,3\}$, which indicates how many members of the team are required to collect the associated reward at each node.
	 %aCOPTW is a newly introduced problem, where no solution exist to compare against the MCW heuristic results. However, as the proposed approach should be evaluated, generated
	 %Small-scale instances were solved by both CPLEX commercial solver and MCW heuristic. %Thereupon, it can provide us with a brighter insight about the solution quality rather than comparison with other heuristic approaches. 
	%experimental setting
	
	In the first study, truncated benchmark sets are designed to solve sufficiently small-size instances by means of both the CPLEX commercial solver and the MCW algorithm. Furthermore, we explored the trade-off between an increased number of available members for service on the one hand and the computation time and objective value on the other hand. We furthermore showed the efficient performance of the proposed heuristic in term of time and accuracy on the large-size designed benchmark instances. All the above computational work was performed on a single node of the National Computational Infrastructure. Each node is equipped with dual 8-core Intel Xeon (Sandy Bridge 2.6 GHz) processors and 32GB of RAM. The algorithm was programmed in C++, using a GCC 5.2.0 compiler. Where applicable, MILP models were solved by the CPLEX 12.6.3 commercial solver in deterministic parallel optimisation mode. All tables show the execution times as elapsed times in seconds.
	%parameter studies
	
	For the parameter studies, after running 584 instances with various parameter settings, we tuned them in a way to define the best possible trade-off between runtime and solution quality. It was decided to change the parameters  within $[0,1.4]$ for $(\lambda, \mu)$ pairs and $[0,3.5]$ for $\vartheta$. Based on the authors' observations the value of $\vartheta$ plays a significant role in the solution quality, thus a broader interval is considered for the newly introduced term in the saving function. Additionally, an incremental size of $0.7$ is large enough to search the feasible region sufficiently and to avoid redundant iterations.
	%results and discussions
	%small instances
	
	For validation and performance evaluation a collection of 456 small-size benchmark instances were generated and solved by means of both CPLEX and the MCW heuristic . A summary of the tests for $10-12$ nodes with $3$ and $4$ team members on instance sets c100, r100 and rc100 is provided in Table 1. The size of truncated instances are chosen in a way to investigate the correlation between the  increase in problem size and exponential growth in computational effort. It is worthwhile to mention that, infeasible edges are excluded in MILP formulations to simplify models for CPLEX.
	
	 In Table 1 computation times are reported for all problems in seconds and the optimality gap is defined by "OPT Gap \%". The average gap from optimal solutions in Table 1 is $2.38\%$ which shows the promising performance of the proposed heuristic. Moreover, it can be seen that the computation time increases drastically with minor changes in the problem size for CPLEX, while that of MCW heuristic experiences negligible change. 
	
	Table 2 gives the summary of results for $24-26$ vertices with the same number of available team members. One can see that CPLEX solves larger problems from the sets c200, r200 and rc200 compared to those in Table 1. This is due the nature of the studied benchmark groups as the time window intervals are different in length and a larger portion of customers can be covered by the same number of team members. In Table 2, the heuristic average computational time remains below one second for all instances, while it takes hours to solve some sets by CPLEX. In Table 2, the average deviation of MCW heuristic from optimal solutions is just $0.45\%$ which is reasonable for a heuristic solution.
	
	To further verify the reliability of MCW, more truncated instances from (pr01-pr10) and (pr11-pr20) sets were tested and results are demonstrated in Table 3 and 4. The proposed algorithm performs similarly in all examined cases which assures its reliability for further runs on larger problems.
	
%	\newpage

		%%%%%%%%%%%%%%%%%%%%%%%%%%%%%%%%%%%%%TABLE 1where p represents number of vehicles

		\begin{table*}[]
			\centering
			\caption{A summary of MCW performance for small-size instances on c100, r100 and rc100 datasets. All computational times are in seconds.}
			
			\begin{tabular}{@{}ccccccccccc@{}}
				\toprule
				\multirow{2}{*}{Set}   & \multirow{2}{*}{\# Vehicles} & \multicolumn{2}{c}{{ 10}} & \multirow{2}{*}{\begin{tabular}[c]{@{}c@{}}Opt\\  Gap \% \end{tabular}} & \multicolumn{2}{c}{{11}} & \multirow{2}{*}{\begin{tabular}[c]{@{}c@{}}Opt\\  Gap \% \end{tabular}} & \multicolumn{2}{c}{{ 12}} & \multirow{2}{*}{\begin{tabular}[c]{@{}c@{}}Opt\\  Gap \% \end{tabular}}  \\ \cmidrule(lr){3-4} \cmidrule(lr){6-7} \cmidrule(lr){9-10}
				&                              & CPLEX         & MCW          &                             & CPLEX      & MCW       &                             & CPLEX        & MCW      &                             \\ \midrule
				\multirow{2}{*}{c100}  & p=3                          & 1.54       & 0.01      & 0.00                        & 0.90        & 0.01     & 0.00                        & 2.82        & 0.10     & 0.00                        \\
				& p=4                          & 1.70       & 0.01      & 0.00                        & 0.75        & 0.01     & 0.00                        & 1.18        & 0.11     & 0.00                        \\
				\multirow{2}{*}{r100}  & p=3                          & 9.23       & 0.17      & 2.67                        & 40.07       & 0.24     & 3.89                        & 1740.62     & 0.36     & 1.19                        \\
				& p=4                          & 12.88      & 0.13      & 3.22                        & 59.92       & 0.19     & 6.74                        & 1455.69     & 0.30     & 5.43                        \\
				\multirow{2}{*}{rc100} & p=3                          & 15.97      & 0.12      & 2.94                        & 43.38       & 0.17     & 2.44                        & 304.96      & 0.24     & 2.20                        \\
				& p=4                          & 8.12       & 0.09      & 1.97                        & 503.42      & 0.13     & 3.52                        & 1908.95     & 0.19     & 6.77                       
				\\ \midrule                     
			\end{tabular}
		\end{table*}
		
		%%%%%%%%%%%%%%%%%%%%%%%%%%%%%%%%%%%%%%%%%%%%%%%%%%%%%%%%%%%%%
		
		%%%%%%%%%%%%%%%%%%%%%%%%%%%%%%%%%%TABLE2
		
		\begin{table*}[]
			\centering
			\caption{A summary of MCW performance for small-size instances on c200, r200 and rc200 dataset. All computational times are in seconds.}
			\label{my-label}
			\begin{tabular}{@{}ccccccccccc@{}}
				\toprule
				\multirow{2}{*}{Set}   & \multirow{2}{*}{\# Vehicles} & \multicolumn{2}{c}{{ 24}} & \multirow{2}{*}{\begin{tabular}[c]{@{}c@{}}Opt\\  Gap \% \end{tabular}} & \multicolumn{2}{c}{{25}} & \multirow{2}{*}{\begin{tabular}[c]{@{}c@{}}Opt\\  Gap \% \end{tabular}} & \multicolumn{2}{c}{{ 26}} & \multirow{2}{*}{\begin{tabular}[c]{@{}c@{}}Opt\\  Gap \% \end{tabular}}  \\ \cmidrule(lr){3-4} \cmidrule(lr){6-7} \cmidrule(lr){9-10}
				&                              & CPLEX         & MCW          &                             & CPLEX      & MCW       &                             & CPLEX        & MCW      &                             \\ \midrule
				\multirow{2}{*}{c200}  & p=3                          & 3.11          & 0.35         & 0.00                        & 4.00       & 0.45      & 0.00                        & 5.44         & 0.59     & 0.00                        \\
				& p=4                          & 2.89          & 0.19         & 0.00                        & 3.93       & 0.34      & 0.00                        & 3.15         & 0.36     & 0.00                        \\
				\multirow{2}{*}{r200}  & p=3                          & 11.84         & 0.48         & 0.74                        & 12.17      & 0.61      & 0.54                        & 19.13        & 0.75     & 0.78                        \\
				& p=4                          & 5.24          & 0.24         & 0.00                        & 6.88       & 0.30    & 0.00                        & 7.71         & 0.37     & 0.00                        \\
				\multirow{2}{*}{rc200} & p=3                          & 22.75         & 0.38         & 0.77                        & 478.37     & 0.52      & 1.20                        & 13585.62     & 0.88     & 3.77                        \\
				& p=4                          & 7.78          & 0.15         & 0.00                        & 8.31       & 0.25      & 0.00                        & 9.83         & 0.37     & 0.45 \\ \midrule                     
			\end{tabular}
		\end{table*}

	%%%%%%%%%%%%%%%%%%%%%%%%%%%%%%%%%%%Table3

	\begin{table*}[]
		\centering
		\caption{A summary of MCW performance for small-size instances on pr01-10 dataset. All computational times are in seconds.}
		\label{my-label}
		\begin{tabular}{@{}ccccccccccc@{}}
			\toprule
			\multirow{2}{*}{Set}   & \multirow{2}{*}{\# Vehicles} & \multicolumn{2}{c}{{ 10}} & \multirow{2}{*}{\begin{tabular}[c]{@{}c@{}}Opt\\  Gap \% \end{tabular}} & \multicolumn{2}{c}{{11}} & \multirow{2}{*}{\begin{tabular}[c]{@{}c@{}}Opt\\  Gap \% \end{tabular}} & \multicolumn{2}{c}{{ 12}} & \multirow{2}{*}{\begin{tabular}[c]{@{}c@{}}Opt\\  Gap \% \end{tabular}}  \\ \cmidrule(lr){3-4} \cmidrule(lr){6-7} \cmidrule(lr){9-10}
			&                              & CPLEX         & MCW          &                             & CPLEX      & MCW       &                             & CPLEX        & MCW      &                             \\ \midrule
			\multirow{2}{*}{pr01-10} & p=3                          & 55.36      & 0.79      & 3.74                        & 84.27      & 0.86     & 2.83                        & 259.63        & 0.97     & 3.43                        \\
			& p=4                          & 64.98      & 0.73      & 3.75                        & 254.00      & 0.81     & 3.16                        & 854.03     & 0.90     & 4.48                        \\ \midrule                     
		\end{tabular}
	\end{table*}

		%%%%%%%%%%%%%%%%%%%%%%%%%%%%%%%%%%%Table4

\begin{table*}[]
	\centering
	\caption{A summary of MCW performance for small-size instances on pr11-20 dataset. All computational times are in seconds.}
	\label{my-label}
	\begin{tabular}{@{}ccccccccccc@{}}
		\toprule
		\multirow{2}{*}{Set}   & \multirow{2}{*}{\# Vehicles} & \multicolumn{2}{c}{{ 19}} & \multirow{2}{*}{\begin{tabular}[c]{@{}c@{}}Opt\\  Gap \% \end{tabular}} & \multicolumn{2}{c}{{20}} & \multirow{2}{*}{\begin{tabular}[c]{@{}c@{}}Opt\\  Gap \% \end{tabular}} & \multicolumn{2}{c}{{ 21}} & \multirow{2}{*}{\begin{tabular}[c]{@{}c@{}}Opt\\  Gap \% \end{tabular}}  \\ \cmidrule(lr){3-4} \cmidrule(lr){6-7} \cmidrule(lr){9-10}
		&                              & CPLEX         & MCW          &                             & CPLEX      & MCW       &                             & CPLEX        & MCW      &                             \\ \midrule
		\multirow{2}{*}{pr11-20} & p=3                          & 10.92      & 0.20      & 3.65                        & 113.72      & 0.31     & 4.93                        & 1497        & 0.43     & 4.97                        \\
		& p=4                          & 19.50      & 0.14      & 1.58                        & 131.95      & 0.25     & 4.43                        & 6917.79     & 0.33     & 4.60                        \\ \midrule                     
	\end{tabular}
\end{table*}

%%%%%%%%%%%%%%%%%%%%%%%%%%%%%%%%%%%%%%%%%%%%%%%%%%%%

%\restylefloat{table}

	%%%%%%%%%%%%%%%%%%%%%%%%%%%%%%%%%%% Table 5

		\begin{table}[]
		    \centering
			\caption{Computational results for the large-size sets}
			\label{my-label}
			\begin{tabular}{@{}cccc@{}}
				\toprule
				Set                      & \# Vehicles & 50    & 100    \\ \midrule
				\multirow{2}{*}{c100}    & p=6         & 21.42 & 138.15 \\
				& p=12        & 10.92 & 153.60 \\
				\multirow{2}{*}{c200}    & p=6         & 3.97  & 131.59     \\
				& p=12        & 3.32  & 44.06     \\
				\multirow{2}{*}{r100}    & p=6         & 26.77 & 148.34 \\
				& p=12        & 22.04 & 184.40 \\
				\multirow{2}{*}{r200}    & p=6         & 10.01 & 190.71     \\
				& p=12        & 7.76  & 86.68    \\
				\multirow{2}{*}{rc100}   & p=6         & 16.25 & 122.84 \\
				& p=12        & 13.13 & 154.61 \\
				\multirow{2}{*}{rc200}   & p=6         & 8.57  & 185.64     \\
				& p=12        & 3.80  & 56.74     \\
				\multirow{2}{*}{pr01-10} & p=6         & 14.85 & 104.89 \\
				& p=12        & 5.13  & 96.72  \\
				\multirow{2}{*}{pr11-20} & p=6         & 38.11 & 274.12     \\
				& p=12        & 14.21 & 246.51     \\ \midrule 
			\end{tabular}
		\end{table}

%%%%%%%%%%%%%%%%%%%%%%%%%%%%%%%%%%%%%%%%%%%%%%%%%%%%%%%%%%%%%%%%%%%%%%%%%%%
 An instance where the MCW achieved an optimal solution on a small set is demonstrated in Fig. 3. For the sake of better presentation, nodes are assigned into the cells of arrays. The  times for starting the service and the corrsponding time window for each succeeding vertex are provided. 
 \begin{figure}[h!]
 	\centering
 	\includegraphics[width=0.46\textwidth,height=0.15\textheight]{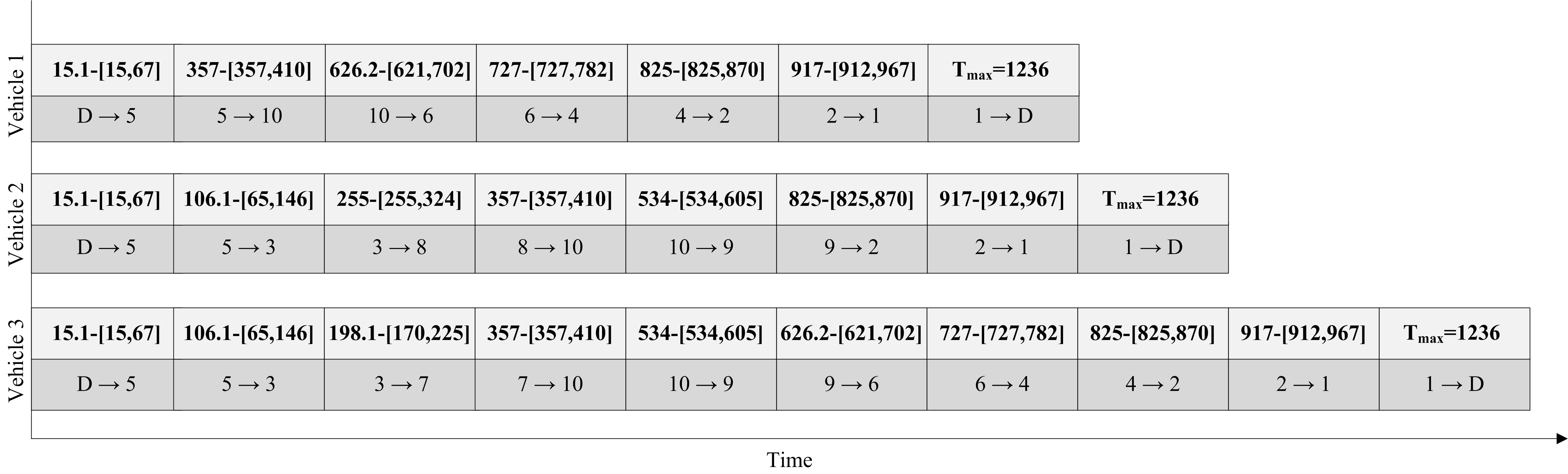}%,height=0.0\textheight
 	\caption{A sample scheduled tour by MCW}
 \end{figure}
 Consider for example, three members leave the depot and arrive at customer number 5 within its time window [15,67], where they start the service simultaneously to collect the associated score at 15.1. After that, second and third members leave customer 5 toward vertex number 3. Finally, all the team members finish their tour by returning to depot before $T_{max}=1236$. One can see that the illustrated routes are highly dependent in a sense that any minor change in orientation of nodes at any tour requires reconstruction of other routes.

 %large-size problems
 The proposed solution method was examined on truncated sets and results were discussed. After the superior performance of the heuristic approach analysed, we carried out further tests on larger instances by using the same approach. Finally, in Table 5 we report a summary of tests for 50 and 100 vertices with 6 and 12 team members. More team members are considered in order to cover a substantial portion of available nodes. Among the (pr01-pr20) test instances those with an insufficient number of nodes are excluded in the reported results. As can be observed, our implementation attains an optimality gap of 2.61\% on instances for which the optimal results are known and solves benchmarks with realistic size within short computational times.

 %Authors had carried out further runs to solve the MILP models after linear relaxation of integer variables (see \cite{lovasz1975ratio}). However, the results revealed that the linear programming relaxation was unable to give us reasonable approximations of the upper bounds, after conducting tests on small instances and comparison with optimal solutions. Therefore, they are not reported in the tables.

	%%%%%%%%%%%%%%%%%%%%%%%%%%%%%%%%%%%%%%%%%%%%%%%%%%%%%%%%%%%%%%%%%%%%%%%	
	
%	\newpage
	\section{Conclusion}
	
	In this paper we investigated the cooperative orienteering problem with time windows, and a new efficient heuristic algorithm is presented to tackle the problem-specific complexities. Although the problem is not newly introduced, we have developed the first heuristic solution to this problem. The proposed algorithm generates optimal solutions for 64.5\% of instances for which the best results are known. In this study, we introduced a new modification to the CW saving heuristic to improve solution quality without losing its simplicity. 
	
	To evaluate the solution approach, a new benchmark set was generated for the COPTW problem. This was achieved by adding the resource requirement attribute to the existing benchmarks (\cite{vansteenwegen2009iterated,montemanni2009ant}) for the TOPTW poblem. Then, the small-sized instances were solved by both the CPLEX solver and the MCW heuristic and results analysed. The optimality gaps and computational times demonstrated the efficiency of the algorithm.
	
	The COPTW is an important class of problem that arises naturally in many applications. For example, in disaster relief aid workers transported by bus must be present at the same locations and time windows as a truck carrying relief supplies so that these supplies can be distributed. As the applications grow, further developments will most likely be neded to represent factors such as soft time windows, and real-time changes to route and capacity constraints. Adding new constraints will require more efficient algorithms where our MCW heuristic at the very least will be able to generate high quality and reliable results that can be used as initial solutions.

	%%%%%%%%%%%%%%%%%%%%%%%%%%%%%%%%%%%%%%%%%%%%%%%%%%%%%%%%%%%%%%%%
	 \section*{Acknowledgement}
	 {
	 	The second author is supported by the Australian Research Council under the Discovery Projects funding scheme (project DP140104246).
	 	
	 	}

	%%%%%%%%%%%%%%%%%%%%%%%%%%%%%%%%%%%%%%%%%%%%%%%%%%%%%%%%%%%%%%%%%%%%%%%%%%%%%%%%%%

  %  \section*{References}
    %plainnat
    %elsarticle-harv

%\bibliography{bibfile}

%\end{multicols}
\end{document}